\def\year{2021}\relax
\documentclass[letterpaper]{article} 
\usepackage{aaai21} 
\usepackage{times} 
\usepackage{helvet} 
\usepackage{courier} 
\usepackage[hyphens]{url} 
\usepackage{graphicx} 
\usepackage{natbib} 
\usepackage{caption} 

\title{Flexible FOND Planning with Explicit Fairness Assumptions$^*$}

\author{
Ivan D. Rodriguez,\textsuperscript{\rm 1} \
Blai Bonet,\textsuperscript{\rm 1} \
Sebastian Sardi\~na,\textsuperscript{\rm 2} \
Hector Geffner\textsuperscript{\rm 1,3}\\
}

\affiliations{
\textsuperscript{\rm 1}Universitat Pompeu Fabra, Barcelona, Spain \\
\textsuperscript{\rm 2}RMIT University, Melbourne, Australia \\
\textsuperscript{\rm 3}{Instituci\'o Catalana de Recerca i Estudis Avan\c{c}ats (ICREA)} \\
ivandaniel.rodriguez01@estudiant.upf.edu, bonetblai@gmail.com, \\
sebastian.sardina@rmit.edu.au, hector.geffner@upf.edu
}

\usepackage{amsmath}
\usepackage{amssymb}
\usepackage{amsthm}
\usepackage{enumerate}
\usepackage{booktabs}
\usepackage[switch]{lineno}  %
\usepackage{xspace}
\usepackage{filecontents}
\usepackage{csvsimple}
\usepackage{etoolbox}
\usepackage{booktabs}
\usepackage{footnote}
\usepackage{footnote}
\usepackage{soul}\setuldepth{x}

\usepackage{pifont}
\newcommand{\cmark}{\ding{51}}%
\newcommand{\xmark}{\ding{55}}%

\usepackage[
    location=inline,    
  ]{moveproofs}

\usepackage{tikz}
\usetikzlibrary{arrows.meta, calc, positioning, backgrounds}

\usepackage{fvextra}
\usepackage{relsize}
\definecolor{darkblue}{rgb}{0.0, 0.0, 0.75}
\definecolor{darkgreen}{rgb}{0.0, 0.70, 0.00}
\fvset{fontfamily=courier,fontsize=\relsize{-1},frame=none,numbers=none,numbersep=4pt,commentchar=!,commandchars=\\\{\}}
\newcommand\atom{\Verb*[fontfamily=courier,fontsize=\relsize{-1}]}

\newcommand{\Omit}[1]{}

\newcommand{\tup}[1]{\langle #1 \rangle}
\newcommand{\pair}[1]{\langle #1 \rangle}

\newcommand{\set}[1]{\ensuremath{\left\{#1 \right\}}}

\newcommand{\citeay}[1]{\citeauthor{#1} (\citeyear{#1})}

\newtheorem{definition}{Definition}

\newtheorem{theorem}[definition]{Theorem}

\newcommand{\CHECK}[1]{\textcolor{red}{\bf *** CHECK: #1 ***}}

\newcommand{\mathtext}[1]{\text{\textit{#1}}}

\newcommand{\G}{\mathcal{G}}

\newcommand{\oneof}{\mathtext{oneof}}

\newcommand{\abst}[2]{\tup{#1;#2}}

\newcommand{\qnpfond}{\textsc{qnp2fond}\xspace}
\newcommand{\fondsat}{\textsc{fond-sat}\xspace}
\newcommand{\prp}{\textsc{prp}\xspace}
\newcommand{\fondasp}{\textsc{fond-asp}\xspace}
\newcommand{\clingo}{\textsc{clingo}\xspace}
\newcommand{\strix}{\textsc{strix}\xspace}

\newcommand{\fond}{FOND\xspace}
\newcommand{\fondplus}{FOND$^+$\xspace}
\newcommand{\sieve}{\textsc{Sieve}\xspace}
\newcommand{\sieveplus}{\textsc{Sieve}$^+$\xspace}

\newcommand{\pplus}{\hspace{-.05em}\raisebox{.15ex}{\footnotesize$\uparrow$}}
\newcommand{\mminus}{\hspace{-.05em}\raisebox{.15ex}{\footnotesize$\downarrow$}}

\newcommand{\GT}[1]{#1{\,>\,}0}
\newcommand{\EQ}[1]{#1{\,=\,}0}

\newcommand{\DEC}[1]{#1\mminus}
\newcommand{\INC}[1]{#1\pplus}

\DeclareMathOperator{\nextltl}{\raisebox{-0.25ex}{\LARGE$\circ$}}
\DeclareMathOperator{\always}{\Box}
\DeclareMathOperator{\eventually}{\Diamond}

\newboolean{proofs}
\setboolean{proofs}{true} 
\newcommand{\Proof}[1]{\ifthenelse{\boolean{proofs}}{\begin{proof}\color{green!50!black} #1 \end{proof}}{}}

\makeatletter
\def\blfootnote{\gdef\@thefnmark{}\@footnotetext}
\makeatother

\begin{document}

\maketitle

\begin{abstract}
  We consider the problem of reaching a propositional goal condition
  in fully-observable non-deterministic (FOND) planning under a general
  class of fairness assumptions that are given explicitly.
  The fairness assumptions  are of the form $A/B$
  and say that state trajectories that contain
  infinite occurrences of an action $a$ from $A$ in a state $s$
  and finite occurrence of actions from $B$,   must also contain
  infinite occurrences of action $a$ in $s$ followed by each one of its possible outcomes.
  The infinite trajectories that violate this condition are deemed as \emph{unfair,}
  and  the solutions   are policies for which \emph{all the fair trajectories reach a goal state.}
  We show that strong and strong-cyclic FOND planning, as well as QNP planning,
  a planning model introduced recently for generalized planning, are all special cases of
  FOND planning with fairness assumptions of this form  which can also be combined.
  \fondplus planning, as this form of planning is called,  combines  the syntax of
  FOND planning with some of the  versatility of LTL for expressing fairness constraints.
  A  new planner is   implemented by reducing \fondplus planning to answer set programs,
  and the performance of the  planner  is evaluated in   comparison with  FOND and QNP planners, and
  LTL synthesis tools.\blfootnote{$^*$This paper is the long version of \cite{ivan:ICAPS21-fond-asp}.}
\end{abstract}

\section{Introduction}


FOND planning is planning with fully observable, non-deterministic state models specified in compact form where a goal state is to be reached~\cite{cimatti:three-models}. In its most common variant, strong-cyclic planning, one is interested in policies that reach states from which the goal can be reached following the policy \cite{CimattiRoverTraverso:AAAI98-strong_cyclic,
DanieleTraversoVardi:RAI99-strong_cyclic}. In another common variant, strong planning~\cite{CimattiRoveriTraverso:AIPS98}, one is interested in policies that reach a goal state in a bounded number of steps. Each form of FOND planning is adequate under a suitable \emph{fairness} assumption; in the case of strong planning, that non-determinism is adversarial (or ``unfair''); in the case of strong-cyclic planning, that non-determinism is fair, in that none of the possible outcomes of a non-deterministic action  can be skipped forever.

FOND planning has become increasingly important as a way of solving other types of problems
such as  \emph{probabilistic (MDP) planning}, where actions have a probabilistic effect
on states \cite{bertsekas:neuro,geffner:book}, \emph{LTL planning}, where
goals to be reached are generalized to temporal conditions that must be satisfied
possibly by plans with cycles \cite{degiacomo:ltl,Camacho.etal:ICAPS19,Aminof.etal:ICAPS19},
and \emph{generalized planning}, where plans are not for single instances but for collections of
instances \cite{srivastava:generalized,hu:generalized}, and they can be obtained
from suitable abstractions encoded as QNP planning problems \cite{Siddharth.etal:AAAI11-QNP,BonetGeffner:JAIR20-QNP}.

A critical limitation of strong, strong-cyclic, and QNP planners, 
is that the fairness assumptions are implicit in their models and solvers,
and as a result, cannot be combined. These combinations, however,
are often needed \cite{camacho:dual,sebastian:dual}, and indeed, 
a recent  FOND planner handles  combinations of fair and adversarial
actions in what is called \emph{Dual FOND planning} \cite{tomas:fond-sat}.
In this work, we go beyond this integration by also enabling the
representation and combination of the conditional fairness assumptions that underlie
QNP planning. This is achieved by extending FOND planning
with a general class of fairness assumptions that are given explicitly as part of the problem.
The fairness assumptions are pairs $A/B$ of sets of actions $A$ and $B$
that say that state trajectories that contain infinite occurrences of actions $a$
from $A$ in a state $s$, and finite occurrences of actions from $B$, must also
contain infinite occurrences of action $a$ in the state $s$
followed by each one of its possible outcomes.
The infinite trajectories that violate this condition are regarded as \emph{unfair.}
The solutions of a FOND problem with {conditional fairness assumptions}
of this type, called a \fondplus problem, are \emph{the policies for which all fair state trajectories reach the goal.}

We show that strong, strong-cyclic, and QNP planning, are all special cases of
\fondplus planning where the fairness assumptions underlying these models can be combined.
\fondplus planning extends  the syntax and semantics of FOND planning
with some of the versatility of the LTL language for expressing fairness constraints.
\Omit{\footnote{
    *** Complexity of \fondplus planning be established with a proof; namely, that it is the same as the complexity of FOND planning,
    although not fully clear how to do that, as compilation of \fondplus into QNPs is not direct. Also, we have to be careful
    about claim that it is a fragment of LTL planning. It is said usually that strong-cyclic planning
    cannot expressed in LTL but CTL*, but this may be about the general goal formula, of reaching the goal from every reachable
    state. A policy that complies with FOND domain and reaches the goal under the LTL fairness
    assumption $(\always \eventually (a \land s) \land \neg (\always \eventually \bigvee_{b \in B})) \supset (\bigwedge_i \always \eventually (a \land s \land s')$
    should work. Hopefully Blai can help with former, and Sebastian with latter.***).}
}
The conditional fairness assumptions $A/B$ correspond to the LTL formulas
$(\always \eventually (s \land a) \land (\neg \always \eventually \bigvee_{b \in B}b)) \supset (\bigwedge_i \always \eventually (a \land s \land \nextltl\! E_i)$,
one for each action $a \in A$, each state $s$, and each possible outcome $E_i$ of the action $a$,
where $s$ stands for the conjunction of literals that $s$ makes true. 
However, unlike LTL synthesis and planning that
are 2EXP-Complete  \cite{ltl:2exp,Camacho.etal:ICAPS19,AminofDeGiacomoRubin:ICAPS20},
\fondplus\ planning is in NEXP (non-deterministic exponential time).

A planner for \fondplus is obtained by reducing \fondplus planning over the explicit state space to an elegant answer set program (ASP), a convenient and high-level alternative to SAT \cite{brewka:asp,vladimir:asp,torsten:asp},
using the facilities provided by the \clingo ASP solver~\cite{torsten:clingo}. The performance of this ASP-based planner is evaluated in comparison with FOND and QNP planners, and LTL synthesis tools.

The paper is organized as follows. We review first strong and strong-cyclic FOND planning, 
and QNP planning. 
We introduce then  \fondplus planning,  where the assumptions underlying these models are stated explicitly and combined,
and present a description of the ASP-based \fondplus planner,  an  empirical evaluation, and a discussion.


\section{FOND Planning}

A FOND model is a tuple $M = \tup{S,s_0,S_G,Act,A,F}$, where $S$ is a finite set of states, $s_0 \in S$ is the initial state, $S_G \subseteq S$ is a non-empty set of goal states, $Act$ is a set of actions, $F(a,s)$ is the set of successor states when action $a$ is executed in state $s$, and $A(s) \subseteq Act$ is the set of actions applicable in state $s$, such that $a \in A(s)$ iff $F(a,s) \not=\emptyset$.
A FOND problem $P$ is a compact description of a FOND model $M(P)$ in terms of a finite set of atoms, so that
the states $s$ in $M(P)$ correspond to truth valuations over the atoms, represented by the set of atoms that are true.
The standard syntax for FOND problems is a simple extension of the STRIPS syntax for classical planning.
A FOND problem is a tuple $P = \tup{At,I,Act,G}$ where $At$ is a set of atoms, $I \subseteq At$ is the
set of atoms true in the initial state $s_0$, $G$ is the set of goal atoms, and $Act$ is a set of actions
with atomic preconditions and effects. If $E_i$ represents the set of positive and negative effects of an action in the classical setting,
action effects in FOND planning can be deterministic of the form $E_i$, or non-deterministic of the form $\oneof(E_1,\ldots,E_n)$.
\Omit{
  Alternatively, a \emph{non-deterministic} action $a$ with effect $oneof(E_1,\ldots,E_n)$ can be regarded as a set of \emph{deterministic}
  actions $b_1$, \ldots, $b_n$ with effects $E_1$, \ldots, $E_n$ respectively, written as $a=\{b_1, \ldots, b_n\}$, all sharing the same preconditions
  of $a$. The application of the non-deterministic action $a$ can be thought as 
  the application of one of the actions $b_i \in a$ chosen non-deterministically.
}

A policy $\pi$ for a FOND problem $P$ is a partial function mapping \emph{non-goal} states into actions.
A policy $\pi$ for $P$ defines a set of, possibly infinite, compatible state trajectories $s_0,s_1,s_2,\ldots$, also called \textbf{$\pi$-trajectories}, where $s_{i+1} \in F(a_i,s_i)$ and $a_i = \pi(s_i)$ for $i \geq 0$.
A trajectory $\tau$ compatible with $\pi$ is \emph{maximal} if it is infinite, or is finite of the form $\tau = s_0,\ldots,s_n$, for some $n \geq 0$, and either $s_n$ is the first state in the sequence being a goal state, $\pi(s_n) \not\in A(s_n)$ (i.e., the action prescribed at $s_n$ is not applicable), or $\pi(s_n) = \bot$ (i.e., no action is prescribed).
Likewise, the policy $\pi$ reaches a state $s$ if there is a $\pi$-trajectory $s_0, \ldots,s_n$ 
where $s=s_n$, and $\pi$ reaches a state $s'$ from a state $s$ if there is a $\pi$-trajectory $s_0, \ldots, s_n$ 
where $s=s_i$ and $s'=s_j$ for $0 \leq i \leq j \leq n$.
A state $s$ is \textbf{recurrent} in  trajectory $\tau$ if it appears an infinite number of times in $\tau$.
The strong and strong-cyclic solutions or policies are usually defined as follows:

\begin{definition}[Solutions]
  A policy $\pi$ is a \textbf{strong solution} for a FOND problem $P$ if all the maximal $\pi$-trajectories reach a goal state, and it is a \textbf{strong-cyclic solution} if $\pi$ reaches a goal state from any state reached by $\pi$.
\end{definition}

The strong solutions correspond also to the strong-cyclic solutions that are acyclic; namely,
where the policies $\pi$ do not give rise to $\pi$-trajectories that can visit a state more than once. 
Alternatively, strong and strong-cyclic solutions can be understood in terms suitable notions of fairness
that establish which $\pi$-trajectories are deemed possible. If we say that \emph{a policy $\pi$ solves problem $P$
when all the \textbf{fair} $\pi$-trajectories reach the goal}, then in strong planning,
all $\pi$-trajectories are deemed fair, while in strong-cyclic planning, all $\pi$-trajectories
are deemed fair \emph{except} those containing a recurrent state $s$
that is followed a finite number of times by a successor $s' \in F(\pi(s),s)$. 

In order to make this alternative ``folk'' characterization of strong and strong-cyclic planning explicit,
let us say that all the actions in strong FOND planning are \textbf{adversarial} (or ``unfair''), and that all the actions in strong-cyclic FOND planning are \textbf{fair}. The state trajectories that are deemed \textbf{fair}
in each setting can then be expressed as follows:

\begin{definition}
  \label{def:standard-fairness}
  If all the actions are \textbf{adversarial}, all $\pi$-trajectories are \textbf{fair}.
  If all the actions are \textbf{fair}, a $\pi$-trajectory $\tau$ is \textbf{fair} iff
  states $s$ that occur an infinite number of times in $\tau$, are followed an infinite number
  of times by each possible successor $s'$ of $s$ given $\pi$, $s' \in F(\pi(s),s)$.
\end{definition}

Provided with these notions of fairness, strong and strong-cyclic solutions can be
characterized equivalently as:

\begin{theorem}\label{thm:fond}
  A policy $\pi$ is a strong (resp.\ strong-cyclic) solution of a FOND problem $P$
  iff all the \textbf{fair} trajectories compatible with $\pi$ in $P$ reach the goal,
  under the assumption that all actions are \textbf{adversarial} (resp. \textbf{fair}).
\end{theorem}
\makeproof{thm:fond}{
  When all actions are adversarial, all trajectories are fair and the claim follows directly.
  For strong-cyclic planning, observe that an infinite \emph{fair} trajectory $\tau$ visits
  each state that is reachable from any recurrent state in $\tau$.
  We show the contrapositive of the two implications. 
  If there is a maximal fair trajectory that does not reach the goal, by the observation, there
  is a state $s$ that is reachable from the initial state and that is not connected to a goal;
  i.e., $\pi$ is not strong-cyclic.
  Conversely, if $\pi$ is not strong-cyclic, there is a state $s$ that is reachable from
  the initial state and that is not connected to the goal, and thus there is a maximal fair
  trajectory that does not reach a goal; i.e., $\pi$ does not solve $P$.
}

Methods for computing strong and strong-cyclic solutions for FOND problems have been developed based on 
OBDDs \cite{cimatti:three-models}, explicit forms of AND/OR search \cite{mynd}, 
classical planners \cite{Muise.etal:ICAPS12-PRP}, and SAT \cite{chatterjee:sat}.
Some of these planners actually handle a \textbf{combination} of {fair} and adversarial actions,
in what is called Dual FOND planning \cite{tomas:fond-sat}.

\section{QNP Planning}

Qualitative numerical planning problems (QNPs) were introduced by \citeay{Siddharth.etal:AAAI11-QNP} as a model for generalized planning, that is, planning for multiple classical instances at once.
QNPs have been used since in other works \cite{bonet:ijcai2017,bonet:aaai2019} and have been analyzed in depth by~\citet{BonetGeffner:JAIR20-QNP}.

The syntax of QNPs is an extension of STRIPS problems $P=\tup{At,I,O,G}$ with negation where $At$ is a set of ground (boolean) atoms, $I$ is a  maximal consistent set of literals from $At$ describing the initial situation, $G$ is a set of literals describing the goal situation, and $O$ is a set of (ground) actions with precondition and effect literals.
A QNP $Q= \tup{At,V,I,O,G}$ extends a STRIPS problem with a set $V$ of \emph{numerical variables} $X$ that can be decremented or incremented \emph{qualitatively}; i.e., by indeterminate positive amounts, without making the variables negative. A numerical variable $X$ can appear in action effects as $\INC{X}$ (increments) and $\DEC{X}$ (decrements),while literals of the form $\EQ{X}$ or $\GT{X}$ (an abbreviation of $X{\neq}0$) can appear everywhere else (initial situation, preconditions, and goals).
The literal \GT{$X}$ is a precondition of all actions with $\DEC{X}$ effects.

A simple example of a QNP is $Q=\tup{At,V,I,O,G}$ with $At=\set{p}$,
$V=\set{n}$, $I=\set{\neg p,\GT{n}}$, $G=\set{\EQ{n}}$, and actions $O=\set{a,b}$ given by
\begin{equation*}\label{eq:ab}
  a= \abst{p, \GT{n}}{\neg p, \DEC{n}} \text{\quad and \quad} b=\abst{\neg p}{p}
\end{equation*}
\noindent where $\abst{C}{E}$ denotes an action with preconditions $C$ and effects $E$.
Thus action $a$ decrements $n$ and negates $p$ that is a precondition of $a$,
and $b$ restores $p$. This QNP represents  an abstraction of the problem of  clearing a block $x$
in Blocksworld instances with stack/unstack actions that include a block $x$.
The numerical variable $n$ stands for the number of blocks above $x$, 
and the boolean variable $p$ stands for the robot gripper being empty.
A policy $\pi$ that solves $Q$ can be expressed by the rules:
\begin{equation*}
  \label{eq:clear}
  \text{\textbf{if} $p$ and $\GT{n}$, \textbf{do} $a$ \quad and \quad \textbf{if} $\neg p \land \GT{n}$, \textbf{do} $b$\,.}
\end{equation*}

A key property of QNPs is that while numerical planning is undecidable \cite{helmert:numeric},
qualitative numerical planning is not. Indeed, a sound and complete, two-step method for solving QNPs was
formulated by~\citet{Siddharth.etal:AAAI11-QNP}:  the QNP $Q$ is converted into a standard FOND problem $P=T_D(Q)$ 
and its (strong-cyclic) solution is checked for termination.
The QNP solutions are in correspondence with the \textbf{strong-cyclic plans}  of the
direct translation $P=T_D(Q)$ that \textbf{terminate.} Moreover, since the number of policies that solve $P$ is finite,
and the termination of each can be verified in finite time, plan existence for QNPs is decidable. 
More recent work has
shown that the complexity of QNP planning is the same as that of FOND planning by introducing
a \textbf{polynomial reduction} from the former into the latter, and another 
in the opposite direction \cite{BonetGeffner:JAIR20-QNP}.

We do not need to get into the formal details of  QNPs but  it  is useful to review the {direct translation} $T_D$ of a QNP $Q$ into a FOND problem $P=T_D(Q)$, and the notion of termination \cite{Siddharth.etal:AAAI11-QNP}. 
Concretely, the translation $T_D$ replaces each numerical variable $n$ by a boolean atom $p_n$ that stands for the (boolean) expression $n=0$. Then, occurrences of the literal $n=0$ in the initial situation, action preconditions, and goals are replaced by $p_n$, while occurrences of the literal $n > 0$ in the same contexts are replaced by $\neg p_n$. 
Likewise, effects $\INC{n}$ are replaced by effects $\neg p_n$, and effects $\DEC{n}$ are replaced by non-deterministic effects $\oneof(p_n,\neg p_n)$.
Actions in the FOND problem $P = T_D(Q)$ with effects $\neg p_n$ (i.e., $n > 0$) are said to ``increment $n$,'' while actions with effects $\oneof(p_n,\neg p_n)$ (i.e., either $n > 0$ or $n = 0$) are said to ``decrement $n$," even if there are no numerical variables in $P$
but just boolean variables. 
This information needs to be preserved in the translation $P=T_D(Q)$, as the semantics of $P$ is not the semantics of FOND problems as assumed by strong or strong-cyclic planners.

\section{Termination and \sieve}

A policy $\pi$ for the FOND problem $P=T_D(Q)$ is said to \textbf{terminate} if all the state trajectories in $P$ that are compatible with the policy $\pi$ and with the fairness assumptions underlying the QNP $Q$, are finite.
Termination is the result of the absence of cycles in the policy that could be  traversed forever.
The latter arises when a cycle includes an action that decrements a numerical variable and none that increments it.
Since numerical variables cannot become negative such cycles eventually terminate.

The procedure called \sieve \cite{Siddharth.etal:AAAI11-QNP} provides a \textbf{sound and complete termination test}
that runs in time that is polynomial in the number of states reached by the policy.
\sieve can be understood as an efficient implementation of the following procedure that operates on a policy graph $G(P,\pi)$ induced
by the FOND problem $P$ and the policy $\pi$, where 
the nodes are the states $s$ that can be reached in $P$ via the policy $\pi$, and
the edges correspond to the state transitions $(s,s')$ that
are possible given the policy $\pi$ (i.e., $s' \in F(\pi(s),s)$). 

Starting with the graph $\G = G(P,\pi)$, \sieve iteratively removes edges from $\G$ until $\G$ becomes acyclic or does not admit further removals.
In each iteration, an edge $(s,s')$ is removed from $\G$ if $\pi(s)$ is an action that decrements a variable $x$ that is not incremented along any path in $\G$ from $s'$ back to $s$.
\sieve \textbf{accepts} the policy $\pi$ iff \sieve\ renders the resulting graph $\G$ acyclic. 
It can be shown that the resulting graph $\G$ is well defined (i.e., it is the same independently of the order in which edges are removed), and that \sieve removes an edge $(s,s')$ when it cannot be  traversed by the policy an infinite number of times.

It is useful to capture the logic of \sieve in terms of an \textbf{inductive definition} that considers states instead of edges:

\begin{definition}[QNP Termination]
  \label{def:qnp:termination}
  Let $\pi$ be a policy for the FOND problem $P=T_D(Q)$ associated with the QNP $Q$.
  The policy $\pi$ \textbf{terminates} in $P$ iff every state $s$ that is reachable by $\pi$ in $P$
  terminates, where a state $s$ \textbf{terminates} iff:\footnote{This inductive definition and
    the ones below imply that there is a \textbf{unique sequence} of state subsets
    $S_0,S_1,\ldots,S_k$ such that $S_{i+1}$ is $S_i$  augmented with all  the states that
    can be added to $S_i$ when assuming that the only  terminating states are those in $S_i$.
    \label{footnote}
  }
  \begin{enumerate}[1.]
    \item there is no cycle on node $s$ (i.e., no path from $s$ to itself),
    \item every cycle on $s$ contains a state $s'$ that \textbf{terminates}, or
    \item $\pi(s)$ decrements a variable $x$, and every cycle on $s$ 
      containing a state $s'$ for which $\pi(s')$ increments $x$, also contains a state $s''$ that \textbf{terminates}.
  \end{enumerate}
\end{definition}


\begin{theorem}\label{thm:qnp:sieve}
  Let $Q$ be a QNPs and $\pi$ a policy.
  Then, \sieve \textbf{accepts} the policy graph $G(P,\pi)$ iff policy $\pi$ \textbf{terminates} in $P$, where $P=T_D(Q)$.
\end{theorem}
\makeproof{thm:qnp:sieve}{
  Since the resulting graph of executing \sieve does not depend on the order in which edges
  are removed, we can consider any execution of \sieve.

  \smallskip\noindent
  \textbf{Forward implication.}
  Let $e_0,e_1,\ldots,e_m$ be the edge removed by \sieve along some execution.
  We show by induction that the state $s_i$ in the edge $Se_i=(s_i,s'_i)$ removed by \sieve is terminating.
  \sieve removes $e_0$ because $\pi(s_0)$ decrements a variable $x$ and there is no cycle involving $s_0$ with actions that increment $x$.
  If so, condition 3 applies, and $s$ must terminate.
  Let us assume that the claim holds for the first $k$ iterations of \sieve, and let us consider the $k+1$-st iteration.
  Edge $e_{k+1}$ is removed because $\pi(s_{k+1})$ decrements a variable $x$ and there is no cycle, in the current graph,  involving $s_{k+1}$ with actions that increment $x$.
  That is, either there is no such cycle in the original graph, or all such cycles have been ``broken'' by the removal of previous edges.
  Hence, by inductive hypothesis, condition 3 applies again and $s$ must terminate.
  Now, if the graph resulting from \sieve is acyclic, all states that have not yet been labeled as terminating
  can be labeled as such using conditions 1 and 2.
  On the other hand, if the resulting graph is not acyclic, then it is easy to show that the
  states in any such cycle cannot labeled as terminating.
  
  \smallskip\noindent
  \textbf{Backward implication.}
  Let $S_0,S_1,\ldots,S_k$ be the state subsets associated with the inductive definition (cf.\ footnote~\ref{footnote}).
  We construct an execution $e_0,e_1,\ldots,e_m$ of \sieve.
  If $s$ in $S_0$, either there is no cycle involving $s$ or $\pi(s)$ decrements a variable that is not incremented along any cycle that involves $s$.
  In the latter case, \sieve removes all edge $(s,s')$.
  Let us assume that we have constructed an execution of \sieve that removes all edges $(s,s')$ for $s$ in $S_0 \cup S_1 \cup \cdots \cup S_i$.
  If $s$ in $S_{i+1}$, either there is no cycle involving $s$ or $\pi(s)$ decrements a variable $x$, and any any cycle that involves $s$ and decrements $x$,
  has a state $s'$ that has been labeled as terminating. Then, using the inductive hypothesis, the current execution of \sieve can
  be extended by the removal of all edge $(s,s')$.
  If all reachable states are labeled as terminating, the constructed execution of \sieve renders an acyclic graph.
  On the other hand, if some reachable state $s$ is not terminating, it can be shown that the resulting graph has
  a cycle that involves $s$; i.e., it is not acyclic.
}

Since solutions to QNPs $Q$ are known to be the strong-cyclic policies of the FOND problem $P=T_D(Q)$ that are accepted by \sieve \cite{Siddharth.etal:AAAI11-QNP,BonetGeffner:JAIR20-QNP}, the solutions for $Q$ can  also be expressed  as:


\begin{theorem}
  \label{thm:qnp}
  A policy $\pi$ is a solution to a QNP $Q$ iff $\pi$ is a strong-cyclic solution of $P=T_D(Q)$ that terminates. 
\end{theorem}
\makeproof{thm:qnp}{
  The result is direct given that $\pi$ is a solution of $Q$ iff
  $\pi$ is a strong-cyclic solution for $P$ accepted by \sieve
  \cite{Siddharth.etal:AAAI11-QNP,BonetGeffner:JAIR20-QNP}, and
  the equivalence between \sieve acceptance and the notion of
  policy termination in Theorem~\ref{thm:qnp:sieve}.
}

The characterization that results from this theorem has been used to verify QNP solutions but not for \emph{computing} them. 
Indeed, the only available complete QNP planner is based on a polynomial reduction of QNP planning into strong-cyclic FOND planning that  avoids the termination test~\cite{BonetGeffner:JAIR20-QNP}.

\section{\fondplus Planning}

In this section, we move from strong, strong-cyclic, and QNP  planning to
the \fondplus setting where  the fairness assumptions underlying these  models can be
explicitly stated and combined. 
A \textbf{\fondplus planning problem} $P_c=\tup{P,C}$ is a FOND problem $P$ extended with a set $C$ of fairness assumptions:

\begin{definition}
  \label{def:fond+:problem}
  A \fondplus problem $P_c=\tup{P,C}$ is a FOND problem $P$ extended with a set $C$ of
  \textbf{(conditional) fairness assumptions} of the form $A_i/B_i$,  $i=1, \ldots, n$ and where each $A_i$ is a set of \textbf{non-deterministic actions} in $P$, and each $B_i$ is a set of actions in $P$  disjoint from  $A_i$.
\end{definition}

The fairness assumptions  play no role in constraining the  state trajectories that are possible by following a policy $\pi$, the so-called $\pi$-trajectories:

\begin{definition}
  \label{def:fond+:trajectory}
  A state trajectory compatible with a policy $\pi$ for the \fondplus problem $P_c=\pair{P,C}$
  is a state trajectory that is compatible with $\pi$ in the FOND problem $P$.
\end{definition}

However, while in strong and strong-cyclic FOND planning all actions are considered  as adversarial and fair, respectively, in the \fondplus setting, each action is labeled fair or unfair depending on the assumptions in $C$ and the trajectory where the action occurs. 
We define what it means for an action $a=\pi(s)$ to behave ``fairly" in a recurrent state $s$ of an infinite $\pi$-trajectory as follows:

\begin{definition}
  \label{def:fond+:fair:action}
  The occurrence of the action $\pi(s)$ in a recurrent state $s$ of a $\pi$-trajectory $\tau$ associated with the \fondplus problem $P_c=\pair{P,C}$
  is \textbf{fair} if for some fairness assumption $A/B \in C$, it is the case that $\pi(s) \in A$ and all the actions in $B$ occur finitely often in $\tau$.
\end{definition}

The meaning of a conditional fairness assumption $A/B$ is that the actions $a \in A$ can be assumed to be \textbf{fair} in any recurrent state $s$ of a $\pi$-trajectory $\tau$, provided that the condition on $B$ holds in $\tau$; namely, that actions in $B$ do \emph{not} occur infinitely often in $\tau$.
Otherwise, if any action in $B$ occurs infinitely often in $\tau$, then $a$ is said to be \textbf{unfair} or \textbf{adversarial}. 
Once actions $\pi(s)$ occurring in recurrent states $s$ are 
``labeled'' in this way, the standard notion of fair trajectories (Definition~\ref{def:standard-fairness}) extends naturally to \fondplus problems:

\begin{definition}\label{def:fond+:fair:path}
  A \textbf{$\pi$-trajectory} $\tau$ for a \fondplus problem $P_c=\pair{P,C}$ is \textbf{fair} if for every recurrent state $s$ in $\tau$ where the action $\pi(s)$ is \textbf{fair} and every 
  possible successor $s'$ of $s$ due to action $\pi(s)$ (i.e., $s' \in F(\pi(s),s)$), state $s$ is immediately followed by state $s'$ in $\tau$ an infinite number of times.
\end{definition}

\Omit{
  Notice that Definitions~\ref{def:fond+:fair:action} and \ref{def:fond+:fair:path} care only about labeling and using the labels of actions
  $a=\pi(s)$ that occur an infinitely number of times in a trajectory compatible with $\pi$. Labeling other actions or other action
  occurrences as fair or unfair is not relevant, as the key question for determining whether a policy $\pi$ solves the \fondplus problem
  $P=\pair{P,C}$ if whether the infinite $\pi$-trajectories, that do not reach the goal, are fair or are not.
}
\noindent The solution of \fondplus problems can then be expressed in a standard way as follows:

\begin{definition}[Solutions]
  \label{def:solution}
  A policy $\pi$ \textbf{solves} the \fondplus problem $P_c=\pair{P,C}$ if the maximal $\pi$-trajectories
  that are \textbf{fair} reach the goal.
\end{definition}

A number of observations can be drawn from these definitions. 
Let us say that one wants to model a non-deterministic action $a$ whose behavior is \textbf{fair} in that it always displays all its possible effects infinitely often in every recurrent state $s$ such that $\pi(s) = a$.
To do so, we consider a fairness constraint $A/B$ in $C$ such that $a \in A$ and $B$ is empty.
On the other hand, to model an \textbf{adversarial} action $b$, one whose behavior is not  fair (may not yield all its effects infinitely often in a recurrent state $s$ with $\pi(s)= b$), we  do not include $b$ in any set $A$.
%
This immediately suggests the way  to capture  standard strong and strong-cyclic planning as special forms of \fondplus planning:

\begin{theorem}\label{thm:strong}
  The \textbf{strong solutions} of a FOND problem $P$ are the solutions
  of the \fondplus problem $P_c=\tup{P,\emptyset}$.
\end{theorem}
\makeproof{thm:strong}{
  Notice that policies are not defined on goal states, and thus there are
  no infinite $\pi$-trajectories that reach a goal, for any policy $\pi$.
  If $\pi$ is a strong solution for $P$, there are no infinite $\pi$-trajectories
  and thus all maximal $\pi$-trajectories are finite and goal reaching;
  i.e., $\pi$ solves $P_c$.

  On the other hand, if $\pi$ solves $P_c$, by definition, the maximal
  $\pi$-trajectories that are fair reach the goal.
  Since there are no constraints, any infinite $\pi$-trajectory is fair and
  non-goal reaching, and thus, if there is any such trajectory, $\pi$ cannot
  solve $P_c$.
  Therefore, all maximal $\pi$-trajectories are finite and goal reaching;
  i.e., $\pi$ is a strong solution for $P$.
}

\begin{theorem}\label{thm:strong-cyclic}
  The \textbf{strong-cyclic solutions} of a FOND problem $P$ are the solutions
  of the \fondplus problem $P_c=\pair{P,\set{A/\emptyset}}$, where $A$ is the
  set of all the non-deterministic actions in $P$.
\end{theorem}
\makeproof{thm:strong-cyclic}{
  Let $\pi$ be a policy for $P$ and let $\tau$ be an infinite $\pi$-trajectory in $P$.
  Since $B$ is empty, any action $\pi(s)$ in $A$ for a recurrent state $s$ in
  $\tau$ is fair (cf.\ Definition~\ref{def:fond+:fair:action}).
  Thus, by Definition~\ref{def:fond+:fair:path} and that $A$ contains all the
  non-deterministic actions, $\tau$ is fair in $P$ iff $\tau$ is fair in $P_c$,
  and by Theorem~\ref{thm:fond}, $\pi$ is a strong-cyclic solution for $P$ iff
  $\pi$ solves $P_c$.
}

Similarly, QNP problems are reduced to \fondplus problems in a direct way, in this case, making
use of both the head $A$ and the condition $B$ in the fairness assumptions $A/B$ in $C$:

\begin{theorem}\label{thm:qnps}
  The solutions of a \textbf{QNP problem} $Q$ are the solutions of the
  \fondplus problem $P_c=\pair{P,C}$ where $P=T_D(Q)$ and $C$ is the set
  of  fairness assumptions $A_i/B_i$, one for each numerical variable $x_i$
  in $Q$, such that $A_i$ contains all the actions in $P$ that decrement $x_i$, 
  and $B_i$ contains all the actions in $P$ that increment $x_i$.
\end{theorem}
\makeproof{thm:qnps}{

  Let $Q$ be a QNP problem, let $P_c=\pair{P,C}$ be the \fondplus problem
  associated with $Q$ where $P=T_D(Q)$, and let $\pi$ be a policy for $Q$ (and $P$).

  \smallskip\noindent
  \textbf{Forward direction.}
  Let us assume that $\pi$ solves $Q$ and suppose it does not solve $P_c$.
  Then, there is a maximal fair (cf.\ Definition~\ref{def:fond+:fair:path}) $\pi$-trajectory $\tau$ that does not reach the goal. 
  The trajectory $\tau$ must be infinite as otherwise $\pi$ would not be strong-cyclic.
  Let $R$ be the set of recurrent states in $\tau$.
  If $s$ in $R$ and the occurrence of $\pi(s)$ is fair, the state $s$ is followed infinitely often in $\tau$ by each $s'\in F(\pi(s),s)$.
  However, the fairness of $\pi(s)$ implies that there is a fairness assumption $A/B$ in $C$ such that $\pi(s) \in A$, and $\pi(s')\in B$ for no $s'\in R$.
  Yet, this implies that the variable $x$ associated with the assumption $A/B$ reaches the value of zero and $\tau$ cannot be infinite.
  Therefore, the occurrence of $\pi(s)$ for any state $s$ in $R$ must be unfair; namely, for any $A/B$ in $C$, if $\pi(s)\in A$ then $\pi(s')\in B$ for some $s'\in R$.
  Since there is at least one state in $R$ where a variable is decremented, no recurrent state in $\tau$ is terminating
  (cf.\ Definition~\ref{def:qnp:termination}), and thus $\pi$ does not solve $Q$.
  Therefore, $\pi$ must solve $P_c$.
  
  \smallskip\noindent
  \textbf{Backward direction.}
  Let us assume that $\pi$ solves $P_c$ and supose it does not solve $Q$.
  Observe that $\pi$ must be strong-cyclic as otherwise it is easy to construct a maximal, non-goal reaching,
  and fair $\pi$-trajectory in $P_c$, contradicting the assumption that $\pi$ solves $P_c$.
  Hence, $\pi$ does not terminate in $Q$; namely, there is an infinite $\pi$-trajectory $\tau$ in $P$
  such that for the set $R$ of its recurrent states, if $\pi(s)$ decrements a variable $x$
  for some $s\in R$, there is another state $s'\in R$ such that $\pi(s')$ increments $x$.
  On the other hand, $R$ must contain a state $s$ such that $\pi(s)$ decrements some variable since such
  actions are the only ones that can generate a cycle in $P$.
  Then, by the definition of the fairness assumptions in $C$, no occurrence of the action $\pi(s)$
  for a state $s$ in $R$ is fair. Thus, $\tau$ is fair in $P_c$ contradicting the assumption
  that $\pi$ solves $P_c$.
  Therefore, $\pi$ must solve $Q$.
}

\section{Example}

By explicitly stating the fairness assumptions underlying strong, strong-cyclic, and QNP planning,
\fondplus planning  integrates these planning models as well. 
We illustrate the new  possibilities with an example.

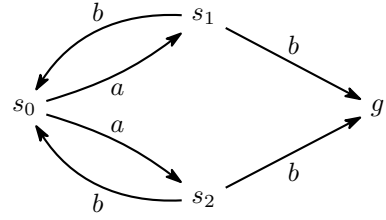
\begin{figure}
  \centering
  \begin{tikzpicture}[thick,>={Stealth[inset=2pt,length=8pt,angle'=33,round]}]
    \node (s0) { $s_0$ };
    \node[above right = 0.75 and 1.80 of s0] (s1) { $s_1$ };
    \node[below right = 0.75 and 1.80 of s0] (s2) { $s_2$ };
    \node[below right = 0.75 and 1.80 of s1] (g) { $g$ };
    \draw[->] (s0) edge[bend right=10] node[below] { $a$ } (s1);
    \draw[->] (s0) edge[bend left=10] node[above] { $a$ } (s2);
    \draw[->] (s1) edge[bend right] node[above] { $b$ } (s0);
    \draw[->] (s2) edge[bend left] node[below] { $b$ } (s0);
    \draw[->] (s1) edge node[above] { $b$ } (g);
    \draw[->] (s2) edge node[below] { $b$ } (g);
  \end{tikzpicture}
  \vskip -.5em
  \caption{Example model for FOND problem $P$ with 4 states, non-deterministic actions $a$ and $b$, and goal state $g$.}
  \label{fig:example}
\end{figure}

Let $P$ be a FOND problem with state set $\set{s_0,s_1,s_2,g}$, two non-deterministic actions $a$ and $b$, initial and goal states being $s_0$ and  $g$, respectively.
Action $a$ can only be applied in state $s_0$, leading to either $s_1$ or $s_2$, whereas action $b$ can be applied only in $s_1$ and $s_2$, leading, in both cases, to either $s_0$ or $g$; see Figure~\ref{fig:example}. 
The FOND problem $P$ admits a single policy, namely, $\pi(s_0)=a$ and $\pi(s_1)=\pi(s_2)=b$,
which we analyze in the context of  different \fondplus problems $P_i = \pair{P,C_i}$ that can be built on top of  $P$ using  different sets of fairness assumptions $C_i$. 
For convenience, in the sets $C_i$, we use $a/b$  to denote the fairness assumption  $\set{a}/\set{b}$, and $a$ to denote the assumption $\set{a}/\emptyset$. 
The marks '\cmark' and '\xmark' express that the policy $\pi$ solves or does not solve, resp., the \fondplus\ problem $P_i$,  where $C_i$ is:
\begin{enumerate}[\xmark]
  \item[\xmark] $C_1 = \set{}$; $a$ and $b$ are adversarial.
  \item[\cmark] $C_2=\set{a,b}$; $a$ and $b$ are fair.
  \item[\xmark] $C_3=\set{a}$; $a$ is fair and $b$ is adversarial.
  \item[\cmark] $C_4=\set{b}$; $b$ is fair and $a$ is adversarial.
  
  \item[\xmark] $C_5=\set{a/b}$; $a$ is conditionally fair on $b$; $b$ adversarial.
  
  \item[\xmark] $C_6=\set{a,b/a}$; QNP like: $a:\DEC{x_1},\INC{x_2}$ and $b:\DEC{x_2}$.
  \item[\cmark] $C_7=\set{b,a/b}$; QNP like: $b:\DEC{x_1},\INC{x_2}$ and $a:\DEC{x_2}$.
  \item[\xmark] $C_8=\set{a/b,b/a}$; QNP like: $a:\DEC{x_1}, \INC{x_2}$ and $b:\DEC{x_2}, \INC{x_1}$.
\end{enumerate}

The subtle cases are the last four. 
The policy $\pi$ does not solve $P_5$ because there are trajectories like $\tau=s_0,s_1,s_0,$ $s_2,s_0,s_1,s_0,\ldots$ that are fair but do not reach the goal. The reason is that while $a/b \in C_5$, the occurrences of the action $a=\pi(s_0)$ in the recurrent state $s_0$ in $\tau$ are not fair. Thus, both $a$ and $b$ have an adversarial semantics in $\tau$.
The policy $\pi$ does not solve $P_6$ either, because in the same trajectory $\tau$, the action $a$ is fair in $s_0$ as $a \in C_6$ but $b$ is not fair in either
$s_1$ or $s_2$, as the assumption $b/a$ is in $C_6$ but $a$ occurs infinitely often in $\tau$.
As a result, $\tau$ is fair but non-goal reaching in $P_6$.
The situation is different in $P_7$, where $b$ is fair and $a$ is unfair. Here, $\tau$ is unfair, as any other trajectory in which some or all the states $s_0$, $s_1$, and $s_2$ occur infinitely often. This is because $b$ being fair in $s_1$ and $s_2$ means that the transitions $(s_1,g)$ and $(s_2,g)$ cannot be skipped forever, and the goal must be reached eventually.
Finally, in $P_8$, the trajectory $\tau$ becomes fair again, as both $a$ and $b$
are adversarial in $\tau$.

\section{Termination and \sieveplus for \fondplus}

We now consider the computation of policies for \fondplus problems.
Initially, we look for a procedure to {verify} if a policy $\pi$ solves a
problem $P_c=\pair{P,C}$, and then transform this \emph{verification procedure}
into a \emph{synthesis procedure}.

The solutions for \fondplus problems are policies that \textbf{terminate in the goal},
a termination condition that combines and goes beyond the solution concept for QNPs
that only requires goal reachability (strong-cyclicity) and termination (finite trajectories).
The termination condition for \fondplus\ planning can be expressed as follows:

\Omit{
  and
  Indeed, the policy above for the  problem  $P_5$ is an example of a policy
  that is strong-cyclic and terminating, but which does not solve $P_5$.
  ..
  But, unlike the definition of termination for QNPs where the notions of termination
  and goal reachability are disentangled, the goal states enter in the definition of
  termination for \fondplus:
}


\begin{definition}[\fondplus termination]
  \label{def:fond+:termination}
  Let $\pi$ be a policy for the \fondplus problem $P_c{=}\pair{P,C}$.
  State $s$ in $P$ \textbf{terminates} iff
  \begin{enumerate}[1.]
    \item $s$ is a goal state,
    \item $s$ is \textbf{fair} and some state $s' \in F(\pi(s),s)$ \textbf{terminates}, or
    \item $s$ is \textbf{not fair}, all states $s' \in F(\pi(s),s)$ \textbf{terminate}, and $F(\pi(s),s)$ is non-empty.
  \end{enumerate}
  where $s$ is \textbf{fair} if for some $A_i/B_i$ in $C$, $\pi(s) \in A_i$, and every path that
  connects $s$ to itself and that contains a state $s'$ with $\pi(s') \in B_i$, also contains a state $s''$ that \textbf{terminates}.
\end{definition}

\fondplus termination expresses a  procedure similar to \sieve, that we call \sieveplus,
that keeps labeling  states $s$ as terminating (the same as removing all  edges from $s$ in the policy graph)
until no states are left or no more states can be labeled. 
The  key difference with \sieve  is that the removals are done backward from the goal as captured
in Definition~\ref{def:fond+:termination}. This is strictly necessary for \sieveplus to be a sound and complete procedure
for \fondplus problems:

\begin{theorem}\label{thm:fond+}
  A policy $\pi$ solves the \fondplus problem $P_c=\pair{P,C}$
  iff all the states $s$ that are reachable by $\pi$ \textbf{terminate}
  according to Definition~\ref{def:fond+:termination}.
\end{theorem}

The solutions to \fondplus problems cannot be  characterized as those of QNPs,  as policies that are strong-cyclic
and terminating in the sense that the policy cannot traverse  edges in the policy graph forever. 
The policy $\pi$ for  the example $P_5$  is indeed strong-cyclic and terminating in this sense,
but as shown above, it does not solve $P_5$. The policy terminates because the action $a$ cannot be done forever, 
but  it does not terminate in a goal state. In QNPs, this cannot happen, as strong-cyclic policies that are terminating,
always terminate in a goal state.


\makeproof{thm:fond+}{
  Let $S_0,S_1,\ldots$ be the chain of state subsets labeled as terminating corresponding
  to Definition~\ref{def:fond+:termination} (cf.\ footnote~\ref{footnote}).
  For a state subset $R$, $\pi(R)$ denotes the set $\{\pi(s):s\in R\}$.

  \smallskip
  \noindent\textbf{Backward implication.} For a proof by contradiction, let us assume
  that every reachable state terminates, and let us suppose that $\pi$ does not solve
  $P_c$; i.e., there is a \emph{maximal and fair} non-goal reaching $\pi$-trajectory $\tau$.
  We consider two cases.

  \smallskip
  \noindent \underline{Case 1:} $\tau=s_0,s_1,\ldots,s_n$ is finite.
  Since $\tau$ is maximal and $s_n$ is not a goal, $F(\pi(s_n),s_n)$ is empty.
  Therefore, $s_n$ cannot terminate contradicting the assumption.

  \smallskip
  \noindent \underline{Case 2:} $\tau$ is infinite.
  Let $R$ be the set of recurrent states in $\tau$, let $j$ be the minimum index
  such that $R\cap S_j\neq\emptyset$, and let $s$ be a state in $R\cap S_j$.
  Clearly, $j>0$ since $S_0$ is the set of goal states and $\tau$ does not reach
  such states. We further consider two subcases:
  \begin{enumerate}[$\bullet$]
    \item $s$ is not fair according to Definition~\ref{def:fond+:termination}.
      Then, every state $s'$ in $F(\pi(s),s)$ must belong to some $S_k$ for $k<j$.
      This is impossible by the choice of $j$.
    \item $s$ is fair according to Definition~\ref{def:fond+:termination}.
      That is, for some constraint $A/B$, $\pi(s)\in A$, and if $\pi(R)\cap B\neq\emptyset$,
      $R$ must contain a state $s''\in S_k$ for $k<j$.
      Since the latter is impossible by the choice of $j$, $\pi(R)\cap B=\emptyset$.
      This implies that the occurrence of the action $\pi(s)$ is fair in the trajectory $\tau$
      (cf.\ Definition~\ref{def:fond+:fair:action}), and thus that $s$ is followed
      in $\tau$ by each of its possible successors $s'\in F(\pi(s),s)$ infinitely
      often; i.e., $F(\pi(s),s)\subseteq R$.
      However, by Definition~\ref{def:fond+:termination}, some such successor $s'$
      must terminate (i.e., $s'\in S_k$ for $k<j$), something that is impossible by
      the choice of $j$.
  \end{enumerate}

  \smallskip
  \noindent\textbf{Forward implication.} For a proof by contradiction, let us assume
  that $\pi$ solves $P_c$, and let us suppose that there is a reachable state $s$ that
  does not terminate. We are going to construct an infinite $\pi$-trajectory $\tau$
  that is fair and does not reach a goal, contradicting the assumption.

  First observe that $s$ is not a goal state and thus it must have successor states.
  If $s$ is fair (cf.\ Definition~\ref{def:fond+:termination}), every state
  $s'\in F(\pi(s),s)$ does not terminate, whereas if $s$ is not fair, some
  state $s'\in F(\pi(s),s)$ does not terminate.
  It is then easy to see that we can construct an infinite $\pi$-trajectory $\tau'$
  seeded at $s$ such that for the set $R$ of its recurrent states:
  \begin{enumerate}[1.]
    \item no state in $R$ terminates,
    \item if $s'$ in $R$ is fair, it is followed infinitely often by each $s''\in F(\pi(s'),s')$, and
    \item if $s'$ in $R$ is not fair, it is followed infinitely often by each $s''\in F(\pi(s'),s')$
      that does not terminate, but there is $s''$ in $F(\pi(s'),s')$ that does not follows $s'$ infinitely often.
  \end{enumerate}
  Since the state $s$ is reachable by $\pi$, we can construct a $\pi$-trajectory $\tau$
  that reaches $s$ from the initial state $s_0$ of $P$ and that then follows $\tau'$.
  The set of recurrent states for $\tau$ is the set $R$ of recurrent state for $\tau'$;
  in particular, $R$ does not contain a goal state. We finish by showing that $\tau$ is fair.

  We do so using Definitions~\ref{def:fond+:fair:action} and \ref{def:fond+:fair:path}
  for fair actions and fair trajectories respectively.
  Let $s'$ be a state in $R$ such that the action $\pi(s')$ is fair in $\tau$;
  in particular, $\pi(s')\in A_i$ for some index $i$.
  If the state $s'$ is fair (cf.\ Definition~\ref{def:fond+:termination}),
  $F(\pi(s'),s)\subseteq R$ by construction of $\tau$.
  Else, if $s'$ is not fair, we show below that the occurrence
  of the action $\pi(s')$ is not fair in $\tau$.
  Hence, in both cases, the trajectory $\tau$ is fair.

  For the last bit, by Definition~\ref{def:fond+:termination},
  if $s'$ is not fair, there is a cycle $\tau''$ that a)~passes over $s'$,
  b)~contains a state $s''$ with $\pi(s'')\in B_i$, and c)~does not contain
  a terminating state.
  Hence, by construction of $\tau$, $R$ contains all the states in $\tau''$.
  However, since there is some state in $F(\pi(s'),s')$ that is not in $R$,
  the occurrence of $\pi(s')$ in $\tau$ is not fair. 
}

\section{\fondplus and Dual FOND Planning}

\fondplus planning subsumes Dual FOND planning where 
fair and adversarial actions can be combined.
In order to show that, let us first recall the latter:

\begin{definition}[\citeauthor{tomas:fond-sat}, \citeyear{tomas:fond-sat}]
  \label{def:dual}
  A Dual FOND problem is a FOND problem where the non-deterministic actions are labeled
  as either \textbf{fair} or \textbf{adversarial}.
  A policy $\pi$ \textbf{solves} a Dual FOND problem $P$ iff
  for all reachable state $s$, $\pi(s){\in}A(s)$, 
  and there is a function $d$ from \textbf{reachable states} into $\{0,\ldots,|S|\}$ such that
  1)~$d(s){=}0$ for goal states, 
  2)~$d(s'){<}d(s)$ for \textbf{some} $s' \in F(\pi(s),s)$ if $\pi(s)$ is fair, and
  3)~$d(s'){<}d(s)$ for \textbf{all} $s' \in F(\pi(s),s)$ if $\pi(s)$ is adversarial.
\end{definition}

For showing that this  semantics coincides with the semantics of
a suitable fragment of \fondplus planning, let us  recast this definition as a termination procedure:

\begin{definition}[Dual FOND termination]
  \label{def:dual:termination}
  Let $\pi$ be a policy for the Dual FOND problem $P$.
  A state $s$ in $P$ \textbf{terminates} iff 
  \begin{enumerate}[1.]
    \item $s$ is a goal state,
    \item $\pi(s)$ is \textbf{fair} and some $s' \in F(\pi(s),s)$ \textbf{terminates}, or
    \item $\pi(s)$ is \textbf{adversarial}, all states $s' \in F(\pi(s),s)$ \textbf{terminate}, and $F(\pi(s),s)$ is non-empty.
  \end{enumerate}
\end{definition}


\begin{theorem}\label{thm:dual}
  $\pi$ is a solution to a Dual FOND problem $P$ iff for every non-goal state $s$
  reachable by $\pi$, $\pi(s) \in A(s)$ and $s$ terminates according to
  Definition~\ref{def:dual:termination}.
\end{theorem}
\makeproof{thm:dual}{
  If $\pi$ is a solution for $P$, there is a minimal function $d$ that
  satisfies Definition~\ref{def:dual}, and that can be used to construct a chain
  $S_0,S_1,\ldots$ of state subsets by $S_i=\{s:\text{$s$ is reachable and $d(s)=i$}\}$.
  It is not difficult to check that such a chain corresponds to the unique chain
  of subsets entailed by Definition~\ref{def:dual:termination} (cf.\ footnote~\ref{footnote}).

  On the other hand, for a policy $\pi$ that entails such a chain $S_0,S_1,\ldots$
  of terminating states that cover all reachable states, the function $d$ on states,
  defined by $d(s)=i$ for the minimum $i$ such that $s\in S_i$,
  satisfies Definition~\ref{def:dual}.
}

The only difference between the termination for Dual FOND and the
one for \fondplus (Def.~\ref{def:fond+:termination}) is that in the former the
fair and adversarial labels are given, while in the latter they are a function
of the explicit fairness assumptions and  policy.
It is easy to show however that Dual FOND problems correspond to the class of  \fondplus
problems with conditional  fairness assumptions $A/B$ with empty $B$:

\begin{theorem}\label{thm:dual:main}
  A policy $\pi$ solves a Dual FOND problem $P'$ iff $\pi$ solves the \fondplus problem $P_c=\pair{P,C}$
  where $P$ is like $P'$ without the action labels, and $C=\{A/B\}$ where $A$ contains all the actions
  labeled as fair in $P'$, and $B$ is empty.
\end{theorem}
\makeproof{thm:dual:main}{
  By Theorems~\ref{thm:dual} and \ref{thm:fond+}, it is enough to
  show that for every $\pi$-reachable and non-goal state $s$ in $P'$, $s$ terminates
  according to Definition~\ref{def:dual:termination} iff $s$ terminates according to
  Definition~\ref{def:fond+:termination}.
  Yet, this is direct since for the unique constraint $A/B$ in $C$, it
  is easy to see that a state $s$ is fair according to Definition~\ref{def:fond+:termination}
  iff $\pi(s)$ is a fair action in $P'$. With this observation, Definition~\ref{def:dual:termination}
  becomes exactly Definition~\ref{def:fond+:termination}.
}

\section{FOND-ASP: An ASP-based \fondplus Planner}

The characterization of \fondplus planning  given in Theorem~\ref{thm:fond+}
allows for  a transparent and direct implementation of a  sound and complete \fondplus planner.
For this,  the planner  \emph{hints} a policy $\pi$ and then each   state reachable by $\pi$
is checked for termination using Definition~\ref{def:fond+:termination}.
The problem of looking for a policy that satisfies this restriction can be expressed
in SAT, although we have found it more convenient to express it as an 
\textbf{answer set program}, a convenient and high-level alternative to SAT \cite{brewka:asp,vladimir:asp,torsten:asp},
using the facilities provided by  \clingo  \cite{torsten:clingo}. 

\Omit{
Exploiting Answer Set Programming (ASP),
a sound and complete hint-and-test procedure is obtained,
where completeness means that the planner eventually tests all possible policies
and if none is found to be solutions, reports that the input problem has no solution.
}

The code for the back-end of the ASP-based \fondplus planner is shown in Figure~\ref{fig:asp}.
The front-end of the planner, not shown, parses an input problem $P_c=\pair{P,C}$
and builds a \emph{flat representation}  of $P_c$ in terms of a number of ground atoms
that are shown in  capitalized predicates  in the figure.
The code in the figure and the facts representing the problem 
are fed to the ASP solver \clingo, which  either returns a (stable) model
for the  program or reports that no such model exists.
In the former case, a policy that solves $P_c$ is  obtained from the atoms
\atom|pi(S,A)| made true by the model.

 
The set of ground atoms providing a flat representation of the problem $P_c$
contains the atoms  \atom|STATE(s)|, \atom|ACTION(a)|, and \atom|TRANSITION(s,a,s')|
for each (reachable) state $s$, ground action $a$ and transition $s' \in F(a,s)$ found
in a reachability analysis from the initial state $s_0$.
In addition, the set  includes the atoms \atom|INITIAL(s0)|, \atom|GOAL(s)| for goal states $s$,
and  \atom|ASET(i,a)| and \atom|BSET(i,b)|  for a fairness assumption $A_i/B_i$ in $C$
if $a \in A_i$ and $b \in B_i$ respectively. 

The program for the \fondplus problem  $P_c$ is denoted as  $T(P_c)$,
while $T(P_c,\pi)$ is used to refer to  the program $T(P_c)$ but with  the line 2 in
Figure~\ref{fig:asp}  replaced by facts \atom|pi(s,a)| when $\pi(s)=a$ for a given policy $\pi$,
and the integrity constraint in line 23  removed. A model $M$ for $T(P_c)$ encodes
a policy $\pi_M$ where $\pi_M(s)=a$ iff \atom|pi(s,a)| holds in $M$.
The formal properties of the \fondasp planner are as follows:


\begin{theorem}\label{thm:fond-asp}
  Let $P_C=\pair{P,C}$ be a \fondplus problem, and let $\pi$ be a policy for $P$.
  Then,
  \begin{enumerate}[1.]
    \item There is a \textbf{unique stable model} $M$ of $T(P_c,\pi)$,
      and $\text{\atom|terminate(s)|}\in M$ iff $s$ terminates (Definition~\ref{def:fond+:termination}).
    \item The policy $\pi$ solves $P_c$ iff the model $M$ for $T(P_c,\pi)$
      satisfies the integrity constraint in line 23 in Figure~\ref{fig:asp}.
    \item $M$ is a model of $T(P_c)$ iff $M$ is the model of $T(P_c,\pi_M)$
      and $M$ satisfies the integrity constraints.
      Thus, \fondasp is a sound and complete planner for \fondplus.
    \item Deciding if $T(P_c,\pi)$ has a model is in P;
      i.e., \fondasp runs in non-deterministic exponential time.
  \end{enumerate}
\end{theorem}
\makeproof{thm:fond-asp}{
  (Sketch.)
  For 1. A model $M$ is stable iff it can be constructed in a ``bottom-up manner''.
  This can be done by constructing a sequence of atom sets $M_0,M_1,\ldots$
  that define the atoms \atom|connected/2|, \atom|blocked/2|, \atom|fair/1|, \atom|terminate/1|,
  and \atom|reachable|. Using induction, it can be shown that any such sequence
  of atom sets lead to the same stable model $M$. This shows uniqueness.
  The existence of $M$ follows since for given $\pi$, there is
  always a stable model $M$ for $T(P_c,\pi)$.
  For 2. By Theorem~\ref{thm:fond+}, $\pi$ solves $P_c$ iff $M$ is the unique model
  of $T(P_c,\pi)$ and $M$ satisfies the integrity constraint.
  For 3. Straightforward by 2.
  For 4. A model for $T(P_c,\pi)$ can be constructed by calculating the terminating
  states for the policy graph for $\pi$. This can be done in time that is polynomial
  in the size of the policy graph which lower bounds the size of $T(P_c)$.
  \fondasp builds $T(P_c)$ in exponential time, then guesses a policy $\pi$ is
  linear time in the size of $T(P_c)$, and finally checks if $T(P_c,\pi)$ has
  a model in time that is polynomial in the size of $T(P_c)$.
}

\begin{figure*}
  \begin{Verbatim}[gobble=1,frame=single,numbers=left,codes={\catcode`$=3},numbersep=-12pt]
!% state(S): S is a state
!% initial(S): S is the initial state
!% goal(S): S is a goal state
!% action(A): A is an action
!% tr(S1,A,S2): S2 belongs to F(S1,A) (F being the transition function)
!% con_A(A,i): action A belongs to set of constraints A_I for index i
!% con_B(A,i): action A belongs to set of constraints B_I for index i
!# show policy/2.
!% policy/2
    \textcolor{darkblue}{% policy, edges, and connectedness}
!   \{ pi(S,A) : ACTION(A) \} = 1 :- STATE(S), not GOAL(S), reachable(S).
    \{ pi(S,A) : ACTION(A) \} = 1 :- STATE(S), not GOAL(S).
!    
!   !% edge/2: policy edges
!   edge(S,T) :- TRANSITION(S,A,T), not GOAL(S), pi(S,A), reachable(S).
    edge(S,T) :- pi(S,A), TRANSITION(S,A,T).
!
!% connected/2: connecting paths given policy
    connected(S,T) :- edge(S,T).
    connected(S,T) :- connected(S,X), edge(X,T), S {\char33}= X.

!% disc/2: disc(s,t) iff there is no st-path, or all such paths have a terminating state
    \textcolor{darkblue}{% blocked(S,T) iff there is no (S,T)-path, or all such paths have a terminating state}
!   blocked(S,T) :- reachable(S), reachable(T), not connected(S,T).
    blocked(S,T) :- STATE(S), STATE(T), not connected(S,T).
    blocked(S,T) :- connected(S,T), terminate(S).
    blocked(S,T) :- connected(S,T), terminate(T).
    blocked(S,T) :- connected(S,T), blocked(X,T) : edge(S,X), connected(X,T).

    \textcolor{darkblue}{% fair(S) iff for some A/B, pi(S) in A and each cycle over S that passes}
    \textcolor{darkblue}{%             over X such that pi(X) in B contains a terminating state}
!   fair(S) :- pi(S,A), ASET(I,A), blocked(X,S) : reachable(X), pi(X,B), BSET(I,B), not blocked(S,X).
    fair(S) :- pi(S,A), ASET(I,A), blocked(X,S) : pi(X,B), BSET(I,B), not blocked(S,X).

    \textcolor{darkblue}{% terminating states}
    terminate(S) :- GOAL(S).
!   terminate(S) :- reachable(S), fair(S), edge(S,T), terminate(T).
!   terminate(S) :- reachable(S), not fair(S), edge(S,_), terminate(T) : edge(S,T).
    terminate(S) :- fair(S), edge(S,T), terminate(T).
    terminate(S) :- not fair(S), edge(S,_), terminate(T) : edge(S,T).

    \textcolor{darkblue}{% reachable states must terminate}
    :- reachable(S), not terminate(S).
    reachable(S) :- INITIAL(S).
!   reachable(S) :- reachable(X), pi(X,A), not GOAL(X), TRANSITION(X,A,S).
    reachable(S) :- reachable(X), not GOAL(X), edge(X,S).
  \end{Verbatim}
  \vskip -1eM
  \caption{\small The concise encoding in \clingo  of the ASP-based \fondplus planner tested (\fondasp).
      The \fondplus problem  enters through the   predicates \atom|STATE/1|, \atom|ACTION/1|, \atom|INITIAL/1|, \atom|GOAL/1|, \atom|TRANSITION/3|,
    \atom|ASET/2| and \atom|BSET/2|, where \atom|ASET(i,A)| (resp.\ \atom|BSET(i,A)|)
    iff action \atom|A| belongs to $A_i$ (resp.\ $B_i$) in the fairness assumption $A_i/B_i$.
    In \clingo; the syntax ``\atom|P : <cond>|''    used in   lines 11, 15, and 20 stands for the implicitly universally
    quantified  conditional ``if \atom|<cond>| then   \atom|P|''. 
  }
  \label{fig:asp}
\end{figure*}

\section{Complexity}

A direct consequence of Theorem~\ref{thm:fond-asp} is that the plan-existence
decision problem for \fondplus is in NEXP (i.e., non-deterministic exponential time).
Since \fond problems are easily reduced to \fondplus problems (Theorem~\ref{thm:strong-cyclic})
and the plan-existence for \fond is EXP-Hard \cite{littman:fond,rintanen:po},
plan-existence for \fondplus is EXP-Hard as well. We conjecture that the NEXP
bound is loose and that plan-existence for \fondplus is EXP-Complete.
In contrast, LTL planning and synthesis is 2EXP-Complete \cite{ltl:2exp}.

\begin{theorem}\label{thm:fond+:complexity}
  The plan-existence problem for \fondplus problems is in NEXP and it is EXP-Hard.
\end{theorem}
\makeproof{thm:fond+:complexity}{
  \textbf{Inclusion.}
  On input problem $P_c=\pair{P,C}$, the state space for $P$ is constructed
  explicitly and a policy $\pi$ for $P_c$ is guessed in non-deterministic
  exponential time. Then, by Theorem~\ref{thm:fond-asp}, the logic program
  $T(P_c,\pi)$ has a model iff $\pi$ solves $P_c$. Deciding whether $T(P_c,\pi)$
  has a model can be done in time polynomial in the size of the program.

  \textbf{Hardness.}
  By Theorem~\ref{thm:strong-cyclic}, A \fond problem can be reduced in polynomial
  time to a \fondplus problem. The claim follows by the EXP-Hardness of the
  plan-existence problem for \fond.
}

\section{Experiments}

We tested \fondasp on three classes of problems:\footnote{Planner available at \url{https://github.com/idrave/aspplanner}}
\fond problems, QNPs, and more expressive \fondplus problems
that do not fit in either class and that can only be addressed
using LTL engines. On each class, we compare \fondasp\ with
the FOND solvers \fondsat \cite{tomas:fond-sat} and \prp \cite{Muise.etal:ICAPS12-PRP},
the QNP solver \qnpfond \cite{BonetGeffner:JAIR20-QNP} using \fondsat and \prp as the underlying FOND solver,
and the LTL-synthesis tool \strix \cite{Meyer.etal:ACTA20-STRIXLTL}.
The pure (strong and strong-cyclic) FOND problems are those in the
FOND-SAT distribution, the QNPs are those by \cite{BonetGeffner:JAIR20-QNP}
and two new families of instances that grow in size  with a parameter.
For more expressive \fondplus\ planning problems, four new families
of problems are introduced that extend the new QNPs with fair and
adversarial actions, with only some being solvable.
The domain and goals of these problems are encoded in LTL in the usual way,
while the fairness assumptions $A/B$ are encoded as described in the introduction.
In all the experiments, time and memory bounds of 1/2 hour and 8GB are enforced.

The results are detailed below. In summary, we observe the following.
For pure FOND benchmarks, \fondasp does not compete with specialized planners like
\prp or \fondsat as these problems span (reachable) state spaces that are just
too large. For QNPs, on the other hand, \fondasp\ does better than \fondsat 
but worse than \prp on the FOND translations.
For expressive \fondplus problems, where these planners cannot be used at all,
\fondasp performs much better than \strix on both solvable and unsolvable problems.

\subsection{FOND Benchmarks}

\fondasp managed to solve a tiny fraction of the benchmarks used for strong and strong-cyclic
planning in the \fondsat distribution. The number of reachable states in these problems
is large (tens of thousands or more) and the size of the grounded ASP program is quadratic in that number.
In general, this seems to limit the scope of  \fondasp to problems with no more than
one thousand states approximately, as suggested by the results in Table~\ref{table:qnp}.
We have observed however that sometimes \fondasp manages 
to solve strong planning problems with more than 100,000 states.
This may have to do with \clingo's grounder or with the state space topology; we do not know the exact reason yet.

\subsection{QNP Problems}


The  two families of QNPs involve the numerical variables
$\{x_i\}_{i=1}^n$ that have all positive values in the initial state.
The goal is to achieve  $x_n=0$. Problems in the \textsc{qnp1} family are solved by means of $n$ sequential
simple loops, while problems in the \textsc{qnp2} family are solved using
 $n$ nested loops.
The actions for problems in \textsc{qnp1} are $b=\abst{\neg p}{p}$, $a_1=\abst{p}{\neg p,\DEC{x_1}}$,
and $a_i=\abst{p,\EQ{x_{i-1}}}{\neg p,\DEC{x_i}}$ for $1< i\leq n$, while those for \textsc{qnp2}
are $b=\abst{\neg p}{p}$,
$a_1=\abst{p}{\neg p, \DEC{x_1}}$, and $a_i=\abst{p,\EQ{x_{i-1}}}{\neg p,\INC{x_{i-1}},\DEC{x_i}}$, $1<i\leq n$.

Table~\ref{table:qnp} shows the results for values of $n$ in $\set{2,3,\ldots,10}$
and different planners, along with the number of reachable states in each problem.
As can be seen, \qnpfond/\prp is the planner that scales best, followed by \fondasp,
\qnpfond/\fondsat, and \strix at the end.
As mentioned, the performance of \fondasp is harmed by a large number of reachable
states. 
While the number of states for the \fond translation produced by \qnpfond
is much larger,  as the translation involves extra propositions, this
number does not necessarily affect the performance of FOND planners like
\fondsat and \prp that can compute compact policies. 
It is also interesting to see how quickly the performance of the LTL engine \strix degrades; it cannot even
solve qnp1-06 which has 14 states. The table also shows results for QNP problems
that capture abstractions for four generalized planning problems, all of which involve small state spaces \cite{BonetGeffner:JAIR20-QNP}.

\begin{table}[t]
  \centering
  \resizebox{\columnwidth}{!}{
    \begin{tabular}{lrr@{}r@{\ \ \ }r@{}rrr}
      \toprule
      &&& \multicolumn{2}{c}{\qnpfond} \\
      \cmidrule{4-5}
      problem  & \#states && \fondsat & \prp && \strix & \fondasp \\
      \midrule
      \begin{filecontents*}{results_qnp1.csv}
problem|states|asp-total-time|fondsat-total-time|prp-total-time|strix-total-time
02| 6|0.00| 0.08|0.09| 3.35
03| 8|0.00| 0.13|0.11| 2.63
04|10|0.00| 0.28|0.14| 5.21
05|12|0.01| 0.60|0.14|98.34
06|14|0.01| 1.27|0.15|  ---
07|16|0.01| 2.54|0.15|  ---
08|18|0.01| 5.54|0.17|  ---
09|20|0.02|12.96|0.21|  ---
10|22|0.02|26.70|0.19|  ---
\end{filecontents*}
\csvreader[separator=pipe,late after line=\\]{results_qnp1.csv}{%
        problem=\problem, states=\states, asp-total-time=\timefondasp, fondsat-total-time=\timefondsat, prp-total-time=\timeprp,strix-total-time=\timestrix}{%
        qnp1-\problem & \states && \timefondsat & \timeprp && \timestrix & \timefondasp}
      \midrule
      \begin{filecontents*}{results_qnp2.csv}
problem|states|asp-total-time|fondsat-total-time|prp-total-time|strix-total-time
02|    8| 0.00|    0.20| 0.18|  2.33
03|   16| 0.01|    1.77| 0.30|  2.31
04|   32| 0.04|   10.00| 0.58| 14.25
05|   64| 0.20|   50.24| 1.15|885.37
06|  128| 1.26|  302.80| 2.53|   ---
07|  256| 7.14|1,969.35| 4.02|   ---
08|  512|54.37|     ---| 6.96|   ---
09|1,024|  ***|     ---|13.22|   ---
10|2,048|  ***|     ---|21.94|   ---
\end{filecontents*}
\csvreader[separator=pipe,late after line=\\]{results_qnp2.csv}{%
        problem=\problem, states=\states, asp-total-time=\timefondasp, fondsat-total-time=\timefondsat, prp-total-time=\timeprp,strix-total-time=\timestrix}{%
        qnp2-\problem & \states && \timefondsat & \timeprp && \timestrix & \timefondasp}
      \midrule
      \begin{filecontents*}{results_qnp_genplan.csv}
problem|states|asp-total-time|fondsat-total-time|prp-total-time|strix-total-time
   Clear| 4|0.00| 0.23|0.13|1.53
      On|16|0.01| 3.69|0.20|3.01
Delivery|12|0.01| 4.07|0.27|1.50
 Gripper|12|0.02|15.43|1.61|2.47
\end{filecontents*}
\csvreader[separator=pipe,late after line=\\]{results_qnp_genplan.csv}{%
        problem=\problem, states=\states, asp-total-time=\timefondasp, fondsat-total-time=\timefondsat, prp-total-time=\timeprp,strix-total-time=\timestrix}{%
        \problem & \states && \timefondsat & \timeprp && \timestrix & \timefondasp}
      \bottomrule
    \end{tabular}
  }
  \vskip -.5eM
  \caption{\small Results for three families of QNPs for \qnpfond paired with the
    \fond solvers \fondsat and \prp, \strix (QNP translated to LTL), and \fondasp.
    Entries '---' and '***' denote solver runs out of time and memory,
    respectively. Time is in seconds.
  }
  \label{table:qnp}
\end{table}

\subsection{More Expressive \fondplus Problems}

The third class of instances consists of four families of problems
obtained from the two QNP families above. The new problems are not ``pure'' QNPs, as they also involve actions with non-deterministic effects on boolean variables that can be adversarial or fair.
Thus, these problems cannot be translated into FOND problems for the use of
planners such as \prp\ or \fondsat.
\Omit{
  The next experiment is to test \fondasp on problems that involve a mixture
  of types of fairness constraints $A_i/B_i$, $A_i$ and $B_i$ non-empty, $B_i$
  empty, and non-deterministic (adversarial) actions that appear in no constraint.
  Starting from the synthetic QNP problems from above, we create two families
  of parametrized problems.
}
For each family \textsc{qnp1} and \textsc{qnp2}, two new families $f01$ and $f11$
of problems are obtained by replacing the action $b=\abst{\neg p}{p}$ 
by the non-deterministic action $b'=\abst{\neg p}{\oneof\set{p,\neg p}}$, leaving the
actions $a_i$ untouched.
Since the action $b'$ does not appear in any fairness assumption, it is adversarial and
thus no problem in the class $f01$ has a solution  as the ``adversary'' may always choose
to leave $p$ false.
The family $f11$ is obtained on top of $f01$ by adding two additional booleans
$q$ and $r$, and two actions $c=\abst{\neg q}{r,\oneof\set{q,\neg q}}$
and $d=\abst{r}{q,\neg r}$ such that: 1)~the actions $a_i$ are modified by adding
$q$ as precondition and $\neg q$ as effect, and 2)~the fairness assumption $A/B$
with $A=\set{b'}$ and $B$ empty is added.
The problems in $f11$ thus involve the QNP-like actions $a_i$, the fair action $b'$,
and the adversarial action $c$, and they all have a solution.

Table~\ref{table:mixture} shows the result for \fondasp and \strix as these
are the only solvers able to handle the combination of fairness assumptions.
As it can be seen, \fondasp scales better than \strix on all of these problems,
the solvable ones (families $f11$) and the unsolvable ones (families $f01$).


\begin{table}[t]
  \centering
  \resizebox{\columnwidth}{!}{
    \begin{tabular}{lr@{} r@{\ \ \ }r@{\ \ \ }r@{\ } r@{\ \ \ } r@{\ \ \ }r@{\ \ \ }r}
      \toprule
      && \multicolumn{3}{c}{$f01$ (unsolvable)} && \multicolumn{3}{c}{$f11$ (solvable)} \\
      \cmidrule{3-5} \cmidrule{7-9}
      problem && \#states & \strix & \fondasp && \#states & \strix & \fondasp \\
      \midrule
      \begin{filecontents*}{results_qnp1_fXX.csv}
problem|f01-states|f01-asp-total-time|f01-strix-total-time|f11-states|f11-asp-total-time|f11-strix-total-time
02| 6|0.00| 3.44|24|0.03| 6.07
03| 8|0.01| 2.42|32|0.04| 6.20
04|10|0.01| 4.13|40|0.07|98.58
05|12|0.01|93.85|48|0.11|  ---
06|14|0.01|  ---|56|0.14|  ---
07|16|0.01|  ---|64|0.19|  ---
08|18|0.01|  ---|72|0.25|  ---
09|20|0.02|  ---|80|0.39|  ---
10|22|0.02|  ---|88|0.34|  ---
\end{filecontents*}
\csvreader[separator=pipe,late after line=\\]{results_qnp1_fXX.csv}{%
        problem=\problem, f01-states=\statesA, f01-asp-total-time=\timefondaspA, f01-strix-total-time=\timestrixA, f11-states=\statesB, f11-asp-total-time=\timefondaspB, f11-strix-total-time=\timestrixB}{%
        qnp1-$f$xx-\problem && \statesA & \timestrixA & \timefondaspA && \statesB & \timestrixB & \timefondaspB}
      \midrule
      \begin{filecontents*}{results_qnp2_fXX.csv}
problem|f01-states|f01-asp-total-time|f01-strix-total-time|f11-states|f11-asp-total-time|f11-strix-total-time
02|    8| 0.00|  3.22|   32| 0.04|  5.85
03|   16| 0.01|  2.25|   64| 0.21|  8.16
04|   32| 0.04| 11.38|  128| 1.55|236.89
05|   64| 0.21|873.09|  256|15.45|   ---
06|  128| 1.25|   ---|  512|46.67|   ---
07|  256|12.13|   ---|1,024|  ***|   ---
08|  512|39.56|   ---|2,048|  ***|   ---
09|1,024|  ***|   ---|4,096|  ***|   ---
10|2,048|  ***|   ---|8,192|  ***|   ---
\end{filecontents*}
\csvreader[separator=pipe,late after line=\\]{results_qnp2_fXX.csv}{%
        problem=\problem, f01-states=\statesA, f01-asp-total-time=\timefondaspA, f01-strix-total-time=\timestrixA, f11-states=\statesB, f11-asp-total-time=\timefondaspB, f11-strix-total-time=\timestrixB}{%
        qnp2-$f$xx-\problem && \statesA & \timestrixA & \timefondaspA && \statesB & \timestrixB & \timefondaspB}
      \bottomrule
    \end{tabular}
  }
  \vskip -.5eM
  \caption{\small Results for four families of \fondplus\ problems
    obtained from the QNPs in Table~1 by playing with the fairness assumptions, some solvable, some unsolvable.
    These problems are handled only by \strix and \fondasp.
    Entries '---' and '***' denote solver runs out of time and memory, respectively.
    Time is in seconds.
  }
  \label{table:mixture}
\end{table}

We finally tested \fondasp over the seven problems considered
in a recent approach to program synthesis over unbounded data structures \cite{Patrizi.etal:KR20}. Although the original specifications are in LTL, these can be all expressed in \fondplus using different types of fairness assumptions. The problems are solved easily by both \fondasp and \strix as their reachable state spaces have very few states.

\Omit{
  \section{Extensions}

  - $n?$, $p?$, non-deterministic policies, ..

  Uncertain effects for QNPs may be already easy to capture in \fondplus model.
  Non-det policies is a true extension, but seems to be easy to capture...

  \textcolor{red}{Another interesting extension is whether something opposite
    to adversaries can be modeled. I.e., ``collaborative'' actions where basically
    the agent ``chooses'' outcome of action. Such action can be used to model
    collaborative agents in a multi-agent setting.
  }
}

\section{Related Work}


The work is related to three threads: SAT-based FOND planning, QNPs, 
and LTL synthesis. The SAT-based FOND planner by \citeay{chatterjee:sat}
expands the state space in full, like \fondasp, but a more recent version
computes compact policies and provides support for Dual FOND planning
\cite{tomas:fond-sat}. 
We have used answer set programs as opposed to CNF encodings exploiting
their high-level modeling language, the natural support for inductive
definitions, and the competitive performance of \clingo \cite{torsten:clingo}.
\fondasp is also a novel QNP planner which can handle non-deterministic effects
on boolean variables.
The formulation actually brings QNP planning into the realm of standard FOND
planning by dealing with the underlying fairness assumptions explicitly.

The use of fairness assumptions connects also to  works on  LTL planning and synthesis
\cite{Camacho.etal:ICAPS19,Aminof.etal:ICAPS19},
and to works addressing  temporally extended goals
\cite{DeGiacomoVardi:ECP99,PatriziLipovetzkyGeffner:IJCAI13,Camacho.etal:AAAI17,Camacho.etal:ICAPS19,AminofDeGiacomoRubin:ICAPS20}.
\Omit{
  Such stream usually considers classical (deterministic) domains or non-deterministic under no
  assumptions (strong solutions) or under the usual fairness on actions' effects (strong-cyclic
  solutions)~\cite{strongly-cyclic,cimatti:three-models}.
}
%
Our work can be seen as a special case of planning under LTL assumptions
 \cite{Aminof.etal:ICAPS19} that targets an LTL  fragment that is relevant for FOND
planning and is computationally simpler. 
While it is possible to express \fondplus tasks as LTL syntheses problems, 
and we  have shown how to do that,
it remains to be seen whether the task can be expressed in a \emph{restricted} LTL fragment that admits more efficient
techniques.
While the \emph{strong} fairness assumption on action effects that is required
cannot be \emph{directly} encoded in GR(1) \cite{Bloem.etal:JCS12-GR1}, 
strong-cyclic FOND planning has been encoded in B\"uchi Games~\cite{DIppolitoRodriguezSardina:JAIR18},
a special case of GR(1). It remains to be investigated whether that encoding can be extended to deal
with \emph{conditional} fairness. 

\section{Summary}

We have formulated an extension of FOND planning that makes
use of explicit fairness assumptions of the form $A/B$ where
$A$ and $B$ are disjoints sets of actions.
While in Dual FOND planning actions are labeled as fair or unfair,
in \fondplus\ planning these labels are a function of the trajectories
and the fairness assumptions: an action $a \in A$ is deemed fair
in a recurrent state if a suitable condition on $B$ holds.
In this way, \fondplus\ generalizes strong, strong-cyclic,
Dual FOND planning, and also QNP planning, which is actually the
only planning setting, excluding LTL planning, that makes use of
the conditions $B$.
We have implemented an effective \fondplus planner by reducing the
problem to answer set programs using \clingo, and evaluated its
performance in relation to FOND and QNP planners, which handle less
expressive problems, and LTL synthesis tools, which handle more
expressive ones. 
We have shown that \fondplus is in NEXP but have not shown yet whether
it is in EXP, like FOND and QNP planning.

\section{Acknowledgments}

We thank the \clingo team for the tool. 
The work is partially supported by an ERC Advanced Grant (No 885107), 
the project TAILOR, funded by an EU Horizon 2020 Grant (No 952215),
and by the Knut and Alice Wallenberg (KAW) Foundation under the WASP program.
H.\ Geffner is a Wallenberg Guest Prof.\ at Link\"oping University, Sweden.

\bibliography{control}

\begin{thebibliography}{38}
\providecommand{\natexlab}[1]{#1}
\providecommand{\url}[1]{\texttt{#1}}
\providecommand{\urlprefix}{URL }
\expandafter\ifx\csname urlstyle\endcsname\relax
  \providecommand{\doi}[1]{doi:\discretionary{}{}{}#1}\else
  \providecommand{\doi}{doi:\discretionary{}{}{}\begingroup
  \urlstyle{rm}\Url}\fi

\bibitem[{Aminof et~al.(2019)Aminof, {De Giacomo}, Murano, and
  Rubin}]{Aminof.etal:ICAPS19}
Aminof, B.; {De Giacomo}, G.; Murano, A.; and Rubin, S. 2019.
\newblock Planning under {LTL} Environment Specifications.
\newblock In \emph{Proc. ICAPS}, 31--39.

\bibitem[{Aminof, {De Giacomo}, and Rubin(2020)}]{AminofDeGiacomoRubin:ICAPS20}
Aminof, B.; {De Giacomo}, G.; and Rubin, S. 2020.
\newblock Stochastic Fairness and Language-Theoretic Fairness in Planning in
  Nondeterministic Domains.
\newblock In \emph{Proc. ICAPS}, 20--28.

\bibitem[{Bertsekas and Tsitsiklis(1996)}]{bertsekas:neuro}
Bertsekas, D.; and Tsitsiklis, J. 1996.
\newblock \emph{Neuro-Dynamic Programming}.
\newblock Athena Scientific.

\bibitem[{Bloem et~al.(2012)Bloem, Jobstmann, Piterman, Pnueli, and
  Sa'ar}]{Bloem.etal:JCS12-GR1}
Bloem, R.; Jobstmann, B.; Piterman, N.; Pnueli, A.; and Sa'ar, Y. 2012.
\newblock Synthesis of Reactive(1) designs.
\newblock \emph{Journal of Computer and System Sciences} 78(3): 911--938.

\bibitem[{Bonet et~al.(2020)Bonet, {De Giacomo}, Geffner, Patrizi, and
  Rubin}]{Patrizi.etal:KR20}
Bonet, B.; {De Giacomo}, G.; Geffner, H.; Patrizi, F.; and Rubin, S. 2020.
\newblock High-Level Programming via Generalized Planning and {LTL} Synthesis.
\newblock In \emph{Proc. KR}, 152--161.

\bibitem[{Bonet et~al.(2017)Bonet, De~Giacomo, Geffner, and
  Rubin}]{bonet:ijcai2017}
Bonet, B.; De~Giacomo, G.; Geffner, H.; and Rubin, S. 2017.
\newblock Generalized Planning: Non-Deterministic Abstractions and Trajectory
  Constraints.
\newblock In \emph{Proc. IJCAI}, 873--879.

\bibitem[{Bonet, Frances, and Geffner(2019)}]{bonet:aaai2019}
Bonet, B.; Frances, G.; and Geffner, H. 2019.
\newblock Learning features and abstract actions for computing generalized
  plans.
\newblock In \emph{Proc. AAAI}, 2703--2710.

\bibitem[{Bonet and Geffner(2020)}]{BonetGeffner:JAIR20-QNP}
Bonet, B.; and Geffner, H. 2020.
\newblock Qualitative Numeric Planning: Reductions and Complexity.
\newblock \emph{Journal of Artificial Intelligence Research ({JAIR})} 69:
  923--961.

\bibitem[{Brewka, Eiter, and Truszczy{\'n}ski(2011)}]{brewka:asp}
Brewka, G.; Eiter, T.; and Truszczy{\'n}ski, M. 2011.
\newblock Answer set programming at a glance.
\newblock \emph{Communications of the ACM} 54(12): 92--103.

\bibitem[{Calvanese, De~Giacomo, and Vardi(2002)}]{degiacomo:ltl}
Calvanese, D.; De~Giacomo, G.; and Vardi, M.~Y. 2002.
\newblock Reasoning about actions and planning in LTL action theories.
\newblock \emph{KR} 2: 593--602.

\bibitem[{Camacho, Bienvenu, and McIlraith(2019)}]{Camacho.etal:ICAPS19}
Camacho, A.; Bienvenu, M.; and McIlraith, S.~A. 2019.
\newblock Towards a Unified View of {AI} Planning and Reactive Synthesis.
\newblock In \emph{Proc. AAAI}, 58--67.

\bibitem[{Camacho and McIlraith(2016)}]{camacho:dual}
Camacho, A.; and McIlraith, S.~A. 2016.
\newblock Strong-Cyclic Planning when Fairness is Not a Valid Assumption.
\newblock In \emph{Proc. KnowProS Workshop}.

\bibitem[{Camacho et~al.(2017)Camacho, Triantafillou, Muise, Baier, and
  McIlraith}]{Camacho.etal:AAAI17}
Camacho, A.; Triantafillou, E.; Muise, C.~J.; Baier, J.~A.; and McIlraith,
  S.~A. 2017.
\newblock Non-Deterministic Planning with Temporally Extended Goals: {LTL} over
  Finite and Infinite Traces.
\newblock In \emph{Proc. AAAI}.

\bibitem[{Chatterjee, Chmel{\'\i}k, and Davies(2016)}]{chatterjee:sat}
Chatterjee, K.; Chmel{\'\i}k, M.; and Davies, J. 2016.
\newblock A symbolic SAT-based algorithm for almost-sure reachability with
  small strategies in POMDPs.
\newblock In \emph{Proc. AAAI}, 3225--3232.

\bibitem[{Cimatti et~al.(2003)Cimatti, Pistore, Roveri, and
  Traverso}]{cimatti:three-models}
Cimatti, A.; Pistore, M.; Roveri, M.; and Traverso, P. 2003.
\newblock Weak, strong, and strong cyclic planning via symbolic model checking.
\newblock \emph{Artificial Intelligence} 147(1-2): 35--84.

\bibitem[{Cimatti, Roveri, and
  Traverso(1998{\natexlab{a}})}]{CimattiRoverTraverso:AAAI98-strong_cyclic}
Cimatti, A.; Roveri, M.; and Traverso, P. 1998{\natexlab{a}}.
\newblock Automatic {OBDD}-based generation of universal plans in
  non-deterministic domains.
\newblock In \emph{Proc. AAAI}, 875--881.

\bibitem[{Cimatti, Roveri, and
  Traverso(1998{\natexlab{b}})}]{CimattiRoveriTraverso:AIPS98}
Cimatti, A.; Roveri, M.; and Traverso, P. 1998{\natexlab{b}}.
\newblock Strong Planning in Non-Deterministic Domains Via Model Checking.
\newblock In \emph{Proc. AIPS}, 36--43.

\bibitem[{Ciolek et~al.(2020)Ciolek, D'Ippolito, Pozanco, and
  Sardi{\~n}a}]{sebastian:dual}
Ciolek, D.; D'Ippolito, N.; Pozanco, A.; and Sardi{\~n}a, S. 2020.
\newblock Multi-Tier Automated Planning for Adaptive Behavior.
\newblock In \emph{Proc. ICAPS}, volume~30, 66--74.

\bibitem[{Daniele, Traverso, and
  Vardi(1999)}]{DanieleTraversoVardi:RAI99-strong_cyclic}
Daniele, M.; Traverso, P.; and Vardi, M.~Y. 1999.
\newblock Strong Cyclic Planning Revisited.
\newblock In \emph{Recent Advances in AI Planning}, 35--48. Springer.

\bibitem[{{De Giacomo} and Vardi(1999)}]{DeGiacomoVardi:ECP99}
{De Giacomo}, G.; and Vardi, M.~Y. 1999.
\newblock Automata-Theoretic Approach to Planning for Temporally Extended
  Goals.
\newblock In \emph{Proc. ECP}, volume 1809 of \emph{LNCS}, 226--238.

\bibitem[{D'Ippolito, Rodr{\'{\i}}guez, and
  Sardi{\~{n}}a(2018)}]{DIppolitoRodriguezSardina:JAIR18}
D'Ippolito, N.; Rodr{\'{\i}}guez, N.; and Sardi{\~{n}}a, S. 2018.
\newblock Fully Observable Non-deterministic Planning as Assumption-Based
  Reactive Synthesis.
\newblock \emph{Journal of Artificial Intelligence Research ({JAIR})} 61:
  593--621.

\bibitem[{Gebser et~al.(2012)Gebser, Kaminski, Kaufmann, and
  Schaub}]{torsten:asp}
Gebser, M.; Kaminski, R.; Kaufmann, B.; and Schaub, T. 2012.
\newblock Answer set solving in practice.
\newblock \emph{Synthesis lectures on artificial intelligence and machine
  learning} 6(3): 1--238.

\bibitem[{Gebser et~al.(2019)Gebser, Kaminski, Kaufmann, and
  Schaub}]{torsten:clingo}
Gebser, M.; Kaminski, R.; Kaufmann, B.; and Schaub, T. 2019.
\newblock Multi-shot ASP solving with clingo.
\newblock \emph{Theory and Practice of Logic Programming} 19(1): 27--82.

\bibitem[{Geffner and Bonet(2013)}]{geffner:book}
Geffner, H.; and Bonet, B. 2013.
\newblock \emph{A Concise Intro. to Models and Methods for Automated Planning}.
\newblock \smash{Morgan-Claypool}.

\bibitem[{Geffner and Geffner(2018)}]{tomas:fond-sat}
Geffner, T.; and Geffner, H. 2018.
\newblock Compact policies for Non-deterministic Fully Observable Planning as
  {SAT}.
\newblock In \emph{Proc. ICAPS}, 88--96.

\bibitem[{Helmert(2002)}]{helmert:numeric}
Helmert, M. 2002.
\newblock Decidability and Undecidability Results for Planning with Numerical
  State Variables.
\newblock In \emph{Proc. AIPS}.

\bibitem[{Hu and {De Giacomo}(2011)}]{hu:generalized}
Hu, Y.; and {De Giacomo}, G. 2011.
\newblock Generalized planning: Synthesizing plans that work for multiple
  environments.
\newblock In \emph{Proc. IJCAI}, 918--923.

\bibitem[{Lifschitz(2019)}]{vladimir:asp}
Lifschitz, V. 2019.
\newblock \emph{Answer set programming}.
\newblock Springer.

\bibitem[{Littman, Goldsmith, and Mundhenk(1998)}]{littman:fond}
Littman, M.~L.; Goldsmith, J.; and Mundhenk, M. 1998.
\newblock The computational complexity of probabilistic planning.
\newblock \emph{Journal of Artificial Intelligence Research} 9: 1--36.

\bibitem[{Luttenberger, Meyer, and Sickert(2020)}]{Meyer.etal:ACTA20-STRIXLTL}
Luttenberger, M.; Meyer, P.~J.; and Sickert, S. 2020.
\newblock Practical synthesis of reactive systems from {LTL} specifications via
  parity games.
\newblock \emph{Acta Informatica} 57(1-2): 3--36.

\bibitem[{Mattm{\"u}ller et~al.(2010)Mattm{\"u}ller, Ortlieb, Helmert, and
  Bercher}]{mynd}
Mattm{\"u}ller, R.; Ortlieb, M.; Helmert, M.; and Bercher, P. 2010.
\newblock Pattern Database Heuristics for Fully Observable Nondeterministic
  Planning.
\newblock In \emph{Proc. ICAPS}, 105--112.

\bibitem[{Muise, McIlraith, and Beck(2012)}]{Muise.etal:ICAPS12-PRP}
Muise, C.; McIlraith, S.~A.; and Beck, J.~C. 2012.
\newblock Improved Non-deterministic Planning by Exploiting State Relevance.
\newblock In \emph{Proc. ICAPS}, 172--180.

\bibitem[{Patrizi, Lipovetzky, and
  Geffner(2013)}]{PatriziLipovetzkyGeffner:IJCAI13}
Patrizi, F.; Lipovetzky, N.; and Geffner, H. 2013.
\newblock Fair {LTL} Synthesis for Non-Determinsistic Systems using Strong
  Cyclic Planners.
\newblock In \emph{Proc. IJCAI}.

\bibitem[{Pnueli and Rosner(1989)}]{ltl:2exp}
Pnueli, A.; and Rosner, R. 1989.
\newblock On the synthesis of an asynchronous reactive module.
\newblock In \emph{ICALP}, 652--671.

\bibitem[{Rintanen(2004)}]{rintanen:po}
Rintanen, J. 2004.
\newblock Complexity of Planning with Partial Observability.
\newblock In \emph{Proc. ICAPS}, 345--354.

\bibitem[{Rodriguez et~al.(2021)Rodriguez, Bonet, Sardi{\~n}a, and
  Geffner}]{ivan:ICAPS21-fond-asp}
Rodriguez, I.~D.; Bonet, B.; Sardi{\~n}a, S.; and Geffner, H. 2021.
\newblock Flexible FOND Planning with Explicit Fairness Assumptions.
\newblock In \emph{Proc. ICAPS}, Accepted.

\bibitem[{Srivastava, Immerman, and Zilberstein(2011)}]{srivastava:generalized}
Srivastava, S.; Immerman, N.; and Zilberstein, S. 2011.
\newblock A new representation and associated algorithms for generalized
  planning.
\newblock \emph{Artificial Intelligence} 175(2): 615--647.

\bibitem[{Srivastava et~al.(2011)Srivastava, Zilberstein, Immerman, and
  Geffner}]{Siddharth.etal:AAAI11-QNP}
Srivastava, S.; Zilberstein, S.; Immerman, N.; and Geffner, H. 2011.
\newblock Qualitative Numeric Planning.
\newblock In \emph{Proc. AAAI}.

\end{thebibliography}

\clearpage

\end{document}